\documentclass{article}

\PassOptionsToPackage{numbers, compress}{natbib}

\usepackage[preprint]{neurips_2023}

\usepackage[utf8]{inputenc} 
\usepackage[T1]{fontenc}    
\usepackage{hyperref}       
\usepackage{url}            
\usepackage{booktabs}       
\usepackage{amsfonts}       
\usepackage{nicefrac}       
\usepackage{microtype}      
\usepackage{xcolor}         

\usepackage{times}
\usepackage{epsfig}
\usepackage{graphicx}
\usepackage{amsmath}
\usepackage{amssymb}
\usepackage{wrapfig}

\usepackage{booktabs}
\usepackage{url}
\usepackage{xcolor}
\usepackage{multirow}
\usepackage{bbding}
\usepackage{makecell}
\usepackage{colortbl} 
\usepackage{float}
\usepackage{arydshln}

\usepackage[noend]{algpseudocode}
\usepackage{algorithmicx,algorithm}

\newtheorem{theorem}{Theorem}

\title{Towards a General Framework for \\ Continual Learning with Pre-training} 

\author{%
  Liyuan Wang\textsuperscript{\rm 1},$\,$ Jingyi Xie\textsuperscript{\rm 1},$\,$ Xingxing Zhang\textsuperscript{\rm 1}\thanks{Corresponding authors.}, $\,$ Hang Su\textsuperscript{\rm 1},$\,$ Jun Zhu\textsuperscript{\rm 1}\footnotemark[1]
  \And 
  \vspace{-.8cm}\\
  Dept. of Comp. Sci. \& Tech., Institute for AI, BNRist Center, THBI Lab, \\Tsinghua-Bosch Joint Center for ML, Tsinghua University, Beijing, China. \\
  \texttt{wly19@tsinghua.org.cn, jingyi\_xie96@163.com} \\
  \texttt{xxzhang1993@gmail.com, \{suhangss, dcszj\}@tsinghua.edu.cn} \\
}

\begin{document}

\maketitle

\begin{abstract}
In this work, we present a general framework for continual learning of sequentially arrived tasks with the use of pre-training, which has emerged as a promising direction for artificial intelligence systems to accommodate real-world dynamics. 
From a theoretical perspective, we decompose its objective into three hierarchical components, including within-task prediction, task-identity inference, and task-adaptive prediction. Then we propose an innovative approach to explicitly optimize these components with parameter-efficient fine-tuning (PEFT) techniques and representation statistics. We empirically demonstrate the superiority and generality of our approach in downstream continual learning, and further explore the applicability of PEFT techniques in upstream continual learning. We also discuss the biological basis of the proposed framework with recent advances in neuroscience. Our code is available at \url{https://github.com/thu-ml/HiDe-Prompt}.
\end{abstract}

\section{Introduction}
To cope with real-world dynamics, continual learning has received widespread attention, especially in the context of pre-training. Through adapting the pre-trained knowledge effectively to downstream tasks, it provides not only positive knowledge transfer but also robustness to catastrophic forgetting \cite{ramasesh2021effect,mehta2021empirical,wang2023comprehensive,zhang2023slca}.
An emerging direction is the implementation of parameter efficient fine-tuning (PEFT) techniques (e.g., Prompt \cite{jia2022visual}, Adapter \cite{rebuffi2017learning}, LoRA \cite{hu2021lora}, FiLM \cite{perez2018film}, etc.), which usually freeze a pre-trained transformer backbone and employ additionally a few parameters to steer representation learning. 
In particular, recent prompt-based approaches \cite{wang2022learning_l2p,wang2022dualprompt,wang2022sprompts,smith2022coda,wang2023hierarchical} focus on \emph{construction} and \emph{inference} of appropriate prompts for each task, and achieve outstanding performance under strong supervised pre-training. However, existing methods usually degrade in performance with challenges in upstream knowledge (e.g., different pre-training paradigms) and downstream tasks (e.g., out-of-distribution and fine granularity), with generality left to be desired.

In this work, we provide an in-depth theoretical analysis of the continual learning objective in the context of pre-training, which can be decomposed into hierarchical components such as \emph{within-task prediction}, \emph{task-identity inference} and \emph{task-adaptive prediction}. 
By leveraging the well-distributed pre-trained representations, we then propose an innovative approach applicable to various PEFT techniques to optimize explicitly the hierarchical components. We perform extensive experiments on downstream continual learning to demonstrate the superiority and generality of our approach, and further explore the applicability of PEFT techniques in upstream continual learning. 
We also provide neurological insights into the proposed framework for acquisition of open-world knowledge.


\section{Hierarchical Decomposition of Continual Learning Objective}
Continual learning aims to master a sequence of tasks represented by their respective training sets $\mathcal{D}_1, ..., \mathcal{D}_T$ and excel on their corresponding test sets. 
Each training set $\mathcal{D}_{t} = \{(\boldsymbol{x}_{t,n}, y_{t,n})\}_{n=1}^{N_t}$, where $|\mathcal{D}_{t}| = N_t$ denotes the size of $\mathcal{D}_t$. $\boldsymbol{x}_{t, n} \in \mathcal{X}_t$ and $y_{t, n} \in \mathcal{Y}_t$ indicate the sample and label elements, respectively. Consider a neural network model with a backbone $f_\theta$ parameterized by $\theta$, and an output layer $h_\psi$ parameterized by $\psi$. This model seeks to learn the projection from $\mathcal{X} = \bigcup_{t=1}^{T} \mathcal{X}_t$ to $\mathcal{Y} = \bigcup_{t=1}^{T} \mathcal{Y}_t$, aiming to predict the label $y = h_\psi(f_\theta(\boldsymbol{x})) \in \mathcal{Y}$ of an unseen test sample $\boldsymbol{x}$ drawn from previous tasks. The backbone function $f_\theta$ is assumed to be pre-trained with a substantial quantity of additional training samples external to each $\mathcal{D}_{t}$.
There are commonly three distinct settings for continual learning \cite{van2019three}: task-, domain-, and class-incremental learning (TIL, DIL, and CIL). Specifically, $\mathcal{Y}_1,...,\mathcal{Y}_T$ are identical for DIL while disjoint for TIL and CIL. The task identity is provided for TIL at test time but is not available for DIL and CIL. 

Here we take CIL as a typical scenario for theoretical analysis, where $\mathcal{Y}_t \cap \mathcal{Y}_{t'} =\emptyset$, $\forall t \neq t'$. 
Let $\mathcal{X}_t=\bigcup_{j}\mathcal{X}_{t,j}$ and $ \mathcal{Y}_t=\{\mathcal{Y}_{t,j}\}$, where $j \in \{1,...,|\mathcal{Y}_t|\}$ indicates the $j$-th class in task $t$.   
Now assume we have a ground event denoted as $\mathcal{D} = \{\mathcal{D}_1, ..., \mathcal{D}_{t}\}$ and a pre-trained model $f_\theta$.
For any sample $\boldsymbol{x} \in \bigcup_{k=1}^{t}\mathcal{X}_{k}$, a general goal of the CIL problem is to learn $P(\boldsymbol{x} \in \mathcal{X}_{i,j}|\mathcal{D},\theta)$, where $i\in\{1,...,t\}$ and $j \in \{1,...,|\mathcal{Y}_i|\}$.
This can be decomposed into two probabilities, including task-identity inference (TII) and within-task prediction (WTP), denoted as 
$P(\boldsymbol{x} \in \mathcal{X}_{i}|\mathcal{D},\theta)$ and
$P(\boldsymbol{x} \in \mathcal{X}_{i,j}|\boldsymbol{x} \in \mathcal{X}_{i},\mathcal{D},\theta)$, respectively.
Based on Bayes' theorem, we have 
\begin{small}
\begin{align}\label{BayesTheorem}
   P(\boldsymbol{x} \in \mathcal{X}_{i,j}|\mathcal{D},\theta) = P(\boldsymbol{x} \in \mathcal{X}_{i,j}|\boldsymbol{x} \in \mathcal{X}_{i},\mathcal{D},\theta) P(\boldsymbol{x} \in \mathcal{X}_{i}|\mathcal{D},\theta).
\end{align}
\end{small}
Let $\bar{i}\in \{1,...,t\}$ and $\bar{j} \in \{1,...,|\mathcal{Y}_i|\}$ be the ground truth of an $\boldsymbol{x}$ w.r.t. the task identity and within-task index. 
Eq.~(\ref{BayesTheorem}) shows that if we can improve either the WTP performance $P(\boldsymbol{x} \in \mathcal{X}_{\bar{i},\bar{j}}|\boldsymbol{x} \in \mathcal{X}_{\bar{i}},\mathcal{D},\theta)$, the TII performance $P(\boldsymbol{x} \in \mathcal{X}_{\bar{i}}|\mathcal{D},\theta)$, or both, then the CIL performance $P(\boldsymbol{x} \in \mathcal{X}_{\bar{i},\bar{j}}|\mathcal{D},\theta)$ would be improved.
However, such an improvement is limited since it is upper-bounded by WTP or TII.
To further improve the CIL performance, we propose a hierarchical decomposition of its objective. 
That is, besides the improvement of $P(\boldsymbol{x} \in \mathcal{X}_{\bar{i},\bar{j}}|\mathcal{D},\theta)$, we also need to improve the performance of task-adaptive prediction (TAP), denoted as $ P(\boldsymbol{x} \in \mathcal{X}^{y}|\mathcal{D},\theta)$, where $\mathcal{X}^{y}$ represents the domain of class $y$ in all previous tasks, and $y=\mathcal{Y}_{\bar{i},\bar{j}}$ is the ground truth label of $\boldsymbol{x}$.
Then the final goal of CIL is formulated as a multi-objective optimization problem, i.e., $\max [P(\boldsymbol{x} \in \mathcal{X}_{\bar{i},\bar{j}}|\mathcal{D},\theta),P(\boldsymbol{x} \in \mathcal{X}^{y}|\mathcal{D},\theta)]$.
Notice that the WTP probability is a categorical distribution over all observed tasks $\{1:t\}$, while the TAP probability is over all observed classes $\bigcup_{k=1}^{t} \mathcal{Y}_k$.

To resolve the problems above, we derive the sufficient and necessary conditions in the context of the widely-used cross-entropy loss.
Specifically, we define 
\begin{small}
\begin{align}\label{H_WTP}
   {H}_{\rm{WTP}}(\boldsymbol{x}) &= \mathcal{H}(\boldsymbol{1}_{\bar{j}},\{P(\boldsymbol{x} \in \mathcal{X}_{\bar{i},j}|\boldsymbol{x} \in \mathcal{X}_{\bar{i}},\mathcal{D},\theta)\}_j),\\
    {H}_{\rm{TII}}(\boldsymbol{x}) &= \mathcal{H}(\boldsymbol{1}_{\bar{i}},\{P(\boldsymbol{x} \in \mathcal{X}_{i}|\mathcal{D},\theta)\}_{i}),\\
   {H}_{\rm{TAP}}(\boldsymbol{x}) &= \mathcal{H}(\boldsymbol{1}_{\bar{c}}, \{P(\boldsymbol{x} \in \mathcal{X}^{c}|\mathcal{D},\theta)\}_{c} ),
\end{align}
\end{small}
where ${H}_{\rm{WTP}}$, ${H}_{\rm{TII}}$, and ${H}_{\rm{TAP}}$ are the cross-entropy values of WTP, TII, and TAP, respectively.
The operation $\mathcal{H}(p,q) \triangleq -\mathbb{E}_{p}[\log q]=-\sum_{i}p_i \log q_i$. $\boldsymbol{1}_{\cdot}$ is a one-hot encoding function.

We now present the first theorem under the CIL scenario: 
\begin{theorem}
    \label{LossError1}
    For continual learning with pre-training, 
    if $ \mathbb{E}_{\boldsymbol{x}} [{H}_{\rm{WTP}}(\boldsymbol{x})] \leq \delta$, $\mathbb{E}_{\boldsymbol{x}} [{H}_{\rm{TII}}(\boldsymbol{x})] \leq \epsilon$, and $\mathbb{E}_{\boldsymbol{x}} [{H}_{\rm{TAP}}(\boldsymbol{x})] \leq \eta$, we have the loss error $\mathcal{L} \in [0, \max\{{\delta +\epsilon},\eta\}]$, regardless whether WTP, TII and TAP are trained together or separately.
\end{theorem}
With the use of cross-entropy, the continual learning performance tends to be better as the bounds are tightened.
In Theorem~\ref{LossError1} we have shown that good performances of WTP, TII and TAP are sufficient to guarantee a good performance of CIL. 
For completeness, we now study the necessary conditions of a well-performed CIL method in Theorem~\ref{LossError2}.
\begin{theorem}
    \label{LossError2}
     For continual learning with pre-training, if the loss error $\mathcal{L}\leq \xi $, then there always exist (1) a WTP, s.t. ${H}_{\rm{WTP}} \leq \xi$; (2) a TII, s.t. ${H}_{\rm{TII}} \leq \xi$; and (3) a TAP, s.t. ${H}_{\rm{TAP}} \leq \xi$.
\end{theorem}
Theorem~\ref{LossError2} suggests that if a continual learning model is well trained (i.e., with low loss), then the WTP, TII and TAP for sequential tasks are always implied to be small.

\section{Optimization of Hierarchical Components}
Motivated by these theoretical insights, we propose to optimize explicitly the hierarchical components (i.e., WTP, TII and TAP) for continual learning with pre-training.
Our proposal stems from two particular advantages of pre-training: (1) the representations can be effectively adapted to downstream tasks through PEFT techniques, and (2) the distributions of unadapted and adapted representations (denoted as $\hat{\mathcal{G}}_c$ and $\mathcal{G}_c$ for each class $c \in \mathcal{Y}_i, i=1,...t-1$, respectively) can be effectively preserved through their statistical information. 
For efficiency and generality, here we employ multiple centroids obtained from K-Nearest Neighbor (KNN) and add Gaussian noise as a specific implementation. 

First, we improve \textbf{WTP} through effectively incorporating task-specific knowledge from each $\mathcal{D}_{t}$. Specifically, we construct task-specific parameters $\boldsymbol{e}_{t}$ with a PEFT technique (e.g., Prompt \cite{jia2022visual}, Adapter \cite{rebuffi2017learning}, LoRA \cite{hu2021lora}, FiLM \cite{perez2018film}, etc.), and optimize ${H}_{\rm{WTP}}$ with cross-entropy (CE).
$\boldsymbol{e}_{1}, ..., \boldsymbol{e}_{t-1}$ are frozen to avoid catastrophic forgetting, while $\boldsymbol{e}_{t}$ is initialized with $\boldsymbol{e}_{t-1}$ to transfer knowledge.
Besides, the adapted representations of $\boldsymbol{e}_{t}$, although allowing the new task to be performed well, may overlap with that of the old tasks and thus affect TAP. 
To overcome this issue, we preserve statistics of adapted representations collected by $f_{\theta}$ and $\boldsymbol{e}_{i}, i=1,...,t-1$, where for classification we calculate the mean $\boldsymbol{\mu}_c$ of $\mathcal{G}_c$ for each class $c \in \mathcal{Y}_i$, and design a \emph{contrastive regularization} (CR):
\begin{small}
\begin{equation}
 \mathcal{L}_{{\rm{CR}}}( \boldsymbol{e}_{t}) =  \sum_{\boldsymbol{h} \in \mathcal{H}_{t}}  
 \frac{1}{\sum_{i=1}^{t-1} |\mathcal{Y}_i|} \sum_{i=1}^{t-1} \sum_{c \in \mathcal{Y}_i} 
 \log \frac{\exp (\boldsymbol{h} \cdot \boldsymbol{\mu}_c / \tau)}
 {\sum_{\boldsymbol{h}' \in \mathcal{H}_{t}} \exp (\boldsymbol{h} \cdot \boldsymbol{h}' / \tau) + \sum_{i=1}^{t-1} \sum_{c \in \mathcal{Y}_i} \exp (\boldsymbol{h} \cdot \boldsymbol{\mu}_c / \tau)},
\end{equation}
\end{small}
where $\mathcal{H}_{t}$ is the embedding transformation of $\mathcal{D}_{t}$ with $f_{\theta}$ and $\boldsymbol{e}_{t}$. $\tau$ is the temperature coefficient, which is insensitive and set to 0.8 in practice. 
Then, the loss function of WTP can be defined as 
\begin{small}
\begin{equation}
\mathcal{L}_{{\rm{WTP}}}( \psi, \boldsymbol{e}_{t}) = \mathcal{L}_{{\rm{CE}}}(\psi, \boldsymbol{e}_{t}) + \lambda \mathcal{L}_{{\rm{CR}}}(\boldsymbol{e}_{t}).
\label{eq.wtp_loss}
\end{equation}
\end{small}
Therefore, the adapted representations of new classes can be well distinguished for WTP while avoiding overlap with the previous ones. $\lambda$ is a hyperparamter to balance the impact of old classes.

Second, we improve \textbf{TII} and \textbf{TAP} through leveraging the approximated distributions of unadapted and adapted representations, respectively.
For ${H}_{\rm{TII}}$, we construct an auxiliary output layer $\hat{h}_{\omega}: \mathbb{R}^D \rightarrow \mathbb{R}^T$ parameterized by $\omega$, learning explicitly the projection from unadapted representations to task identity via cross-entropy:
\begin{small}
\begin{equation}
\mathcal{L}_{{\rm{TII}}}(\omega) = \frac{1}{\sum_{i=1}^{t} |\mathcal{Y}_i|} \sum_{i=1}^{t} \sum_{c \in \mathcal{Y}_i} \sum_{\hat{\boldsymbol{h}} \in \hat{\mathcal{H}}_{i,c}} - \log \frac{\exp(\hat{h}_{\omega}(\hat{\boldsymbol{h}})[i])}{\sum_{j=1}^{t} \exp(\hat{h}_{\omega}(\hat{\boldsymbol{h}})[j])},
\label{eq.tii_loss}
\end{equation}
\end{small}
where $\hat{\mathcal{H}}_{i,c}$ is constructed by sampling an equal number of pseudo representations from $\hat{\mathcal{G}}_c$ for $c \in \mathcal{Y}_i$ and $i=1,...,t$. 
Similarly, the final output layer $h_\psi: \mathbb{R}^D \rightarrow \mathbb{R}^{|\mathcal{Y}|}$ is further optimized for ${H}_{\rm{TAP}}$: 
\begin{small}
\begin{equation}
\mathcal{L}_{{\rm{TAP}}}(\psi) = \frac{1}{\sum_{i=1}^{t} |\mathcal{Y}_i|} \sum_{i=1}^{t} 
\sum_{c \in \mathcal{Y}_i} \sum_{\boldsymbol{h} \in \mathcal{H}_{i,c}} - \log \frac{\exp(h_{\psi}(\boldsymbol{h})[c])}{\sum_{j=1}^{t}\sum_{c' \in \mathcal{Y}_j} \exp(h_{\psi}(\boldsymbol{h})[c'])}, 
\label{eq.tap_loss}
\end{equation}
\end{small}
where $\mathcal{H}_{i,c}$ is constructed by sampling an equal number of pseudo representations from $\mathcal{G}_c$ for $c \in \mathcal{Y}_i$ and $i=1,...,t$. 
As $\omega$ and $\psi$ are usually \emph{light-weight}, the optimization of TII and TAP is computationally efficient. At test time, our approach predicts the task identity $i = \hat{h}_{\omega}(f_\theta(\boldsymbol{x}))$ and then the label $y = h_\psi(f_\theta(\boldsymbol{x}; \boldsymbol{e}_{i}))$.


\section{Experiment}

\textbf{Experimental Setup:} We consider two CIL benchmarks that are widely used for downstream continual learning \cite{wang2022learning_l2p,wang2022dualprompt,smith2022coda}, such as Split ImageNet-R \cite{krizhevsky2009learning_cifar} of 200-class natural images and Split CUB-200 \cite{wah2011caltech} of 200-class bird images, randomly split into 10 incremental tasks. After learning multiple incremental tasks, we further evaluate upstream continual learning with the ability of few-shot learning, i.e., adapting the backbone to a N-way K-shot task \cite{finn2017model} randomly sampled from subsequent unseen classes. 
We consider supervised and self-supervised pre-training on ImageNet-21K, denoted as Sup-21K and iBOT-21K, respectively.

\textbf{Experimental Result:} We implement two representative PEFT techniques as the task-specific parameters in our approach, such as Prompt \cite{jia2022visual} (adjusting intermediate inputs through prepending a short sequence of learnable prompt parameters) and LoRA \cite{hu2021lora} (adjusting backbone parameters through adding a learnable low-rank parameter matrix). 
We first evaluate the performance of \emph{downstream continual learning} with different pre-training paradigms and CIL benchmarks. As shown in Table~\ref{table:downstream}, the performance of state-of-the-art prompt-based approaches degrades remarkably under self-supervised pre-training (e.g., iBOT-21K) and fine-grained classification (e.g., Split CUB-200), while both versions of our approach outperform them significantly. 

On the other hand, a potential limitation of prompt-based methodologies is that, the pre-trained knowledge in backbone parameters \emph{cannot} be updated and enriched from incremental tasks, which has been rarely discussed in previous literature.
Motivated by this, we then consider \emph{upstream continual learning}, i.e., the ability of accumulating knowledge in backbone parameters. Specifically, after downstream continual learning of multiple incremental tasks, we evaluate the performance of the backbone to perform few-shot learning of an additional task randomly sampled from subsequent unseen classes. As shown in Table~\ref{table:upstream}, the backbone adapted by the LoRA version of our approach acquires strong improvements in few-shot learning, compared to the unadapted backbone of the Prompt version. In addition to Split ImageNet-R and Split CUB-200 that split all tasks from the same dataset, we further consider a mixture of tasks sampled from CUB-200 \cite{wah2011caltech} and Cars-196 \cite{krause20133d} datasets, where the improvements remain significant. These results demonstrate the importance and feasibility of synchronizing upstream and downstream continual learning.

\newcommand{\tabincell}[2]{\begin{tabular}{@{}#1@{}}#2\end{tabular}}
\begin{table*}[t]
	\centering
    \vspace{-0.6cm}
    \caption{Performance of downstream continual learning. 
    PTM: pre-trained model. FAA: final average accuracy. CAA: cumulative average accuracy. FFM: final forgetting measure. 
    } 
	\smallskip
      \renewcommand\arraystretch{1.25}
    \small{
	\resizebox{0.85\textwidth}{!}{ 
	\begin{tabular}{c|l|ccc|ccc}
	 \hline
        \multirow{2}{*}{PTM} & \multirow{2}{*}{\,\,\,\,\,\,\,\,\,\,\,\, Method} & \multicolumn{3}{c|}{Split ImageNet-R} & \multicolumn{3}{c}{Split CUB-200} \\
        & & FAA ($\uparrow$) & CAA ($\uparrow$) & FFM ($\downarrow$) & FAA ($\uparrow$) & CAA ($\uparrow$) & FFM ($\downarrow$)\\
        \hline
       \multirow{7}*{\tabincell{c}{Sup-21K}} 
       &L2P \cite{wang2022learning_l2p} &63.65 &67.25 &7.51 &75.58 &80.32 &6.38 \\ 
       &DualPrompt \cite{wang2022dualprompt} &68.79 &71.96 &4.49 &81.32 &83.45 &5.31 \\ 
       &S-Prompt \cite{wang2022sprompts} &69.68 &72.50 &3.29 &81.51 &83.24 &4.48 \\
       &CODA-Prompt \cite{smith2022coda} &70.03 &74.26 &5.17 &74.34 &80.71 &7.42 \\
       \cdashline{2-8}[2pt/2pt]
       &Ours-Prompt &73.55 &75.93 &0.95 &84.60 &83.87 &0.21 \\ 
       &Ours-LoRA &69.59 &74.18 &8.68 &85.26 &86.56 &3.58 \\ 
       &Ours-Adapter &70.48 &75.03 &7.89 &84.69 &86.51 &4.10 \\
        \hline
       \multirow{7}*{\tabincell{c}{iBOT-21K}} 
       &L2P \cite{wang2022learning_l2p} &55.35 &58.62 &3.73 &45.93 &56.02 &9.20 \\ 
       &DualPrompt \cite{wang2022dualprompt} &54.55 &58.69 &5.38 &41.46 &54.57 &14.03 \\ 
       &S-Prompt \cite{wang2022sprompts} &55.16 &58.48 &4.07 &39.88 &53.71 &13.15 \\ 
       &CODA-Prompt \cite{smith2022coda} &61.22 & 66.76 &9.66 &47.79 &59.24 &11.81 \\
       \cdashline{2-8}[2pt/2pt]
       &Ours-Prompt &70.63 &72.94 &1.31 &72.27 &73.66 &1.94 \\ 
       &Ours-LoRA &70.94 &74.92 &5.61 &71.75 &76.57 &5.33 \\ 
       &Ours-Adapter &72.07 &76.01 &5.48 &74.29 &78.74 &5.35 \\
       \hline
	\end{tabular}
	} }
	\label{table:downstream}
	\vspace{-0.2cm}
\end{table*}

\begin{table*}[t]
	\centering
    \vspace{-0.2cm}
    \caption{Performance of upstream continual learning. After learning 8 incremental tasks, we present the accuracy of learning each N-way K-shot (NWKS) task sampled from subsequent classes.} 
	\smallskip
      \renewcommand\arraystretch{1.25}
    \small{
	\resizebox{0.88\textwidth}{!}{ 
	\begin{tabular}{c|l|cc|cc|cc}
	 \hline
        \multirow{2}{*}{PTM} & \multirow{2}{*}{\,\,\,\,\,Method} & \multicolumn{2}{c|}{Split ImageNet-R} & \multicolumn{2}{c|}{Split CUB-200} & \multicolumn{2}{c}{Split CUB \& Cars}  \\
        & & 5W1S ($\uparrow$) & 5W5S ($\uparrow$) & 5W1S ($\uparrow$) & 5W5S ($\uparrow$) & 5W1S ($\uparrow$) & 5W5S ($\uparrow$)\\
        \hline
       \multirow{2}*{\tabincell{c}{Sup-21K}} 
       &Ours-Prompt &49.88 &67.88 &77.50 &80.63 &65.87 &82.13 \\ 
       &Ours-LoRA &67.00 &82.88 &79.50 &92.88 &71.87 &86.93 \\ 
        \hline
       \multirow{2}*{\tabincell{c}{iBOT-21K}} 
       &Ours-Prompt &42.87 &64.63 &36.50 &62.88 &34.87 &58.93 \\ 
       &Ours-LoRA &58.38 &78.87 &53.63 &79.75 &40.60 &68.40 \\ 
       \hline
	\end{tabular}
	} }
	\label{table:upstream}
	\vspace{-0.5cm}
\end{table*}

\section{Discussion}
In this work, we propose a general framework for continual learning in the context of pre-training, with decomposing the objective into three hierarchical components (i.e., WTP, TII and TAP) and optimizing them explicitly with PEFT techniques and representation statistics.
Through extensive experiments, we demonstrate the superiority and generality of our approach in downstream continual learning, and further elaborate on the importance of upstream continual learning, which requires updating the backbone parameters rather than instructing only (intermediate) inputs. 
Interestingly, the proposed framework requires sequential invocation of the unadapted and (task-specific) adapted representations for inference, which is consistent with recent advances in biological learning and memory \cite{lei2022social,lei2022adult} that the activation of non-memory cells and memory cells (as well as their specific sub-populations) is internally switched. This connection potentially bridges the intrinsic mechanisms of biological and artificial intelligence in acquisition of open-world knowledge. 


\clearpage

\section*{Acknowledgements}
This work was supported by the National Key Research and Development Program of China (No. 2020AAA0106302), NSFC Projects (Nos.~62061136001, 92248303, 62106123, 61972224), BNRist (BNR2022RC01006), Tsinghua Institute for Guo Qiang, and the High Performance Computing Center, Tsinghua University. L.W. is also supported by Shuimu Tsinghua Scholar, and J.Z. is also supported by the XPlorer Prize.

\bibliographystyle{plain}
\bibliography{egbib}

\end{document}